\documentclass[twoside,11pt]{article}

\usepackage{jmlr2e}
\usepackage{mathtools}
\usepackage[all,cmtip]{xy}
\usepackage{amsmath}

\usepackage{lastpage}
\usepackage{pythonhighlight}

\newcommand{\eg}{e.\,g., }
\newcommand{\ie}{i.\,e. }

\firstpageno{1}

\begin{document}

\title{GraKeL: A Graph Kernel Library in Python}

\author{\name Giannis Siglidis \email yiannis.siglidis@lip6.fr\\
        \addr LIP6, UPMC Universit\'e Paris 6, Sorbonne Universit\'es\\
        Paris, France\\
        \name Giannis Nikolentzos \email nikolentzos@lix.polytechnique.fr \\
        \name Stratis Limnios \email stratis.limnios@polytechnique.edu \\
        \name Christos Giatsidis \email giatsidis@lix.polytechnique.fr\\
        \name Konstantinos Skianis \email kskianis@lix.polytechnique.fr\\
        \addr LIX, \'Ecole Polytechnique\\
        Palaiseau, France\\
        \name Michalis Vazirgiannis \email mvazirg@lix.polytechnique.fr\\
        \addr LIX, \'Ecole Polytechnique\\
        Palaiseau, France\\
        and\\
        Department of Informatics, Athens University of Economics and Business\\
        Athens, Greece
}


\maketitle

\begin{abstract}
The problem of accurately measuring the similarity between graphs is at the core of many applications in a variety of disciplines.
Graph kernels have recently emerged as a promising approach to this problem.
There are now many kernels, each focusing on different structural aspects of graphs.
Here, we present GraKeL, a library that unifies several graph kernels into a common framework.
The library is written in Python and adheres to the scikit-learn interface.
It is simple to use and can be naturally combined with scikit-learn's modules to build a complete machine learning pipeline for tasks such as graph classification and clustering.
The code is BSD licensed and is available at: \url{https://github.com/ysig/GraKeL}.
\end{abstract}


\section{Introduction}
In recent years, graph-structured data has experienced an unprecedented growth in many domains, ranging from social networks to bioinformatics.
Several problems of increasing interest involving graphs call for the use of machine learning techniques.
Measuring the similarity or distance between graphs is a key component in many of those machine learning algorithms.
Graph kernels have emerged as an effective tool for tackling the graph similarity problem.
A graph kernel is a function that corresponds to an inner-product in a Hilbert space, and can be thought of as a similarity measure defined directly on graphs.
The main advantage of graph kernels is that they allow a large family of machine learning algorithms, called kernel methods, to be applied directly to graphs.

GraKeL is a package that provides implementations of several graph kernels.
The library is BSD licensed, and is publicly available on a GitHub repository encouraging collaborative work inside the machine learning community.
The library is also compatible with scikit-learn, a standard package for performing machine learning tasks in Python \citep{pedregosa2011scikit}.
Given scikit-learn's current inability to handle graph-structured data, the proposed library was built on top of one of its templates, and can serve as a useful tool for performing graph mining tasks.
At the same time, it enjoys the overall object-oriented syntax and semantics defined by scikit-learn.
Note that graphs are combinatorial structures and lack the convenient mathematical context of vector spaces.
Hence, algorithms defined on graphs exhibit increased diversity compared to the ones defined on feature vectors.
Therefore, bringing together all these kernels under a common framework is a challenging task, and the main design decisions behind GraKeL are presented in the following sections.

\section{Underlying Technologies}
Inside the Python ecosystem, there exist several packages that allow efficient numerical and scientific computation.
GraKeL relies on the following technologies for implementing the currently supported graph kernels:\\
$\bullet$ NumPy: a package that offers all the necessary data structures for graph representation.
Furthermore, it offers numerous linear algebra operations serving as a fundamental tool for achieving fast kernel calculation \citep{walt2011numpy}.\\
$\bullet$ SciPy: Python's main scientific library.
It contains a large number of modules, ranging from optimization to signal processing.
Of special interest to us is the support of sparse matrix representations and operations \citep{virtanen2020scipy}.\\
$\bullet$ Cython: allows the embedding of C code in Python.
It is used to address efficiency issues related to non-compiled code in high-level interpreted languages such as Python, as well as for integrating low-level implementations \citep{behnel2011cython}.\\
$\bullet$ scikit-learn: a machine learning library for Python.
It forms the cornerstone of GraKeL since it provides the template for developing graph kernels.
GraKeL can also interoperate with scikit-learn for performing machine learning tasks on graphs \citep{pedregosa2011scikit}.\\
$\bullet$ BLISS: a tool for computing automorphism groups and canonical labelings of graphs.
It is used for checking graph isomorphism between small graphs \citep{JunttilaKaski:ALENEX2007}.\\
$\bullet$ CVXOPT (optional): a package for convex optimization in Python.
It is used for solving the semidefinite programming formulation that computes the Lov\'asz number $\vartheta$ of a graph \citep{andersen2013cvxopt}.

\section{Code Design}
In GraKeL, all graph kernels are required to inherit the \texttt{Kernel} class which inherits from the scikit-learn's \texttt{TransformerMixin} class and implements the following four methods:\\
1. \texttt{fit}: Extracts kernel dependent features from an input graph collection.\vspace{0.1cm}\\
2. \texttt{fit\_transform}: Fits and calculates the kernel matrix of an input graph collection.\\
3. \texttt{transform}: Calculates the kernel matrix between a new  collection of graphs and the one \nobreak\hskip1pt given as input to \texttt{fit}.\\
4. \texttt{diagonal}: Returns the self-kernel values of all the graphs given as input to \texttt{fit} along with those given as input to \texttt{transform}, provided that this method has been called.
This method is used for normalizing kernel matrices.

All kernels are unified under a submodule named \texttt{kernels}.
They are all wrapped in a general class called \texttt{GraphKernel} which also inherits from scikit-learn's \texttt{TransformerMixin}.
Besides providing a unified interface, it is also useful for applying other operations such as the the Nystr{\"o}m method, while it also facilitates the use of kernel frameworks that are currently supported by GraKeL.
Frameworks like the Weisfeiler Lehman algorithm \citep{shervashidze2011weisfeiler} can use any instance of the \texttt{Kernel} class as their base kernel.

The input is required to be an \texttt{Iterable} collection of graph representations.
Each graph can be either an \texttt{Iterable} consisting of a graph representation object (\eg adjacency matrix, edge dictionary), vertex attributes and edge attributes or a \texttt{Graph} class instance.
The vertex and edge attributes can be discrete (a.k.a. vertex and edge labels in the literature of graph kernels) or continuous-valued feature vectors.
Note that some kernels cannot handle vector attributes, while others assume unlabeled graphs.
Furthermore, through its \texttt{datasets} submodule, GraKeL facilitates the application of graph kernels to several popular graph classification datasets contained in a public repository \citep{KKMMN2016}.

\section{Comparison to Other Software}
In the past years, researchers in the field of graph kernels have made available small collections of graph kernels.
These kernels are written in various languages such as Matlab and Python, and do not share a general common structure that would provide an ease for usability.
In the absence of software packages to compute graph kernels, the \texttt{graphkernels} library was recently developed \citep{sugiyama2017graphkernels}.
All kernels are implemented in C++, while the library provides wrappers to R and Python.
The above packages and the \texttt{graphkernels} library exhibit limited flexibility since kernels are not wrapped in a meaningful manner and their implementation does not follow object-oriented concepts.
GraKeL, on the other hand, is a library that employs object-oriented design principles encouraging researchers and developers to integrate their own kernels into it.

Moreover, the \texttt{graphkernels} library contains only a handful of kernels, while several state-of-the-art kernels are missing.
On the other hand, GraKeL provides implementations of a larger number of kernels.
In a quick comparison, the \texttt{graphkernels} library provides variations of 5 kernels and 1 kernel framework, while GraKeL provides implementations of 15 kernels and 2 kernel frameworks.
Moreover, GraKeL is compatible with the scikit-learn pipeline allowing easy and fast integration inside machine learning algorithms.
In addition, given the diversities in the evaluation of machine learning methods, GraKeL provides a common ground for comparing existing kernels against newly designed ones.
This can be of great interest to researchers trying to evaluate kernels they have come up with.
It should also be mentioned that GraKeL is accompanied by detailed documentation including several examples of how to apply graph kernels to real-world data.

Furthermore, even though GraKeL is implemented in Python, as shown in Figure~\ref{fig:runtime_comparison} below, several of its kernels are more efficient than the corresponding implementations in \texttt{graphkernels}.
Due to space limitations, we only present the results for a single benchmark dataset (\ie ENZYMES).
The rest of the results can be found in the documentation\footnote{\url{https://ysig.github.io/GraKeL/latest/benchmarks/comparison.html}}.

\begin{figure}
  \centering
  \includegraphics[width=\linewidth]{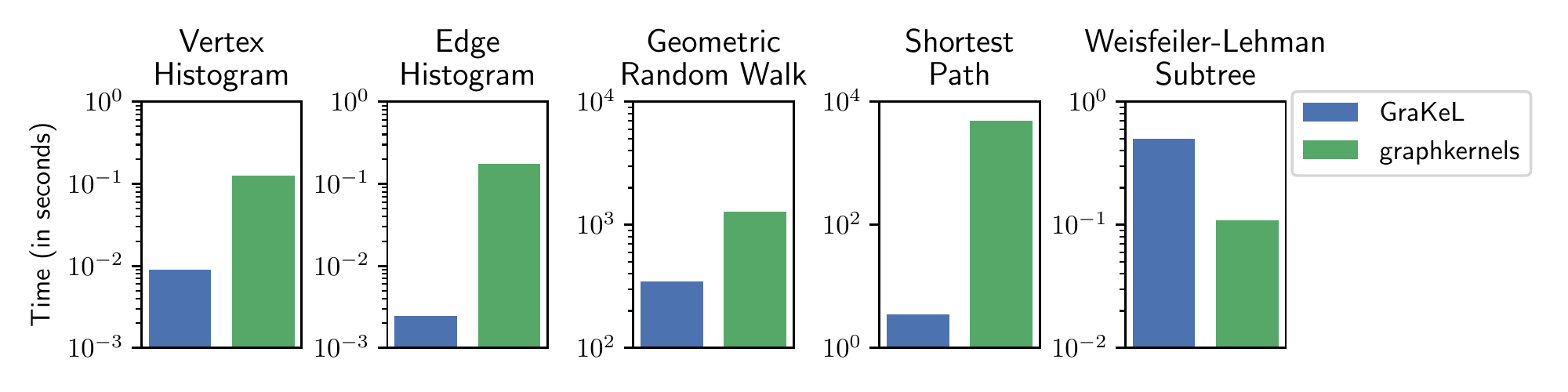}
  \caption{Running time (in seconds) for kernel computation on the ENZYMES dataset using the GraKeL and \texttt{graphkernels} libraries.}
  \label{fig:runtime_comparison}
\end{figure}

\section{Sample Code}
The most common use of a graph kernel is the one where given a collection of training graphs $\mathcal{G}_n$ (of size $n$) and a collection of test graphs $\mathcal{G}_m$ (of size $m$), the goal is to compute two separate kernel matrices: ($1$) an $n \times n$ matrix between all the graphs of $\mathcal{G}_n$, and ($2$) a $m \times n$ matrix between the graphs of $\mathcal{G}_m$ and those of $\mathcal{G}_n$.
This can be accomplished by running the \texttt{fit\_transform} method on $\mathcal{G}_n$, and then the \texttt{transform} method on $\mathcal{G}_m$.
Then, these matrices can be passed on to the SVM classifier to perform graph classification.
The following example demonstrates the use of GraKeL for performing graph classification on a standard dataset.
\begin{python}
>>> from grakel.datasets import fetch_dataset
>>> from sklearn.model_selection import train_test_split
>>> from grakel.kernels import ShortestPath
>>> from sklearn.svm import SVC
>>> from sklearn.metrics import accuracy_score
>>>
>>> MUTAG = fetch_dataset("MUTAG", verbose=False)
>>> G, y = MUTAG.data, MUTAG.target
>>> G_train, G_test, y_train, y_test = train_test_split(G, y, test_size=0.1, random_state=42)
>>>
>>> sp_kernel = ShortestPath()
>>> K_train = sp_kernel.fit_transform(G_train)
>>> K_test = sp_kernel.transform(G_test)
>>>
>>> clf = SVC(kernel='precomputed').fit(K_train, y_train)
>>> y_pred = clf.predict(K_test)
>>> print("accuracy: 
accuracy: 84.21 %
\end{python}

\section{Conclusion}
GraKeL is a library that implements several state-of-the-art graph kernels, while remaining user-friendly.
It relies on the scikit-learn's pipeline, and it can thus be easily integrated into various machine learning applications.

\acks{This work was supported by the Labex DigiCosme ``Grakel'' project.}


\vskip 0.2in
\bibliography{biblio}

\end{document}